\documentclass[conference]{IEEEtran}
\IEEEoverridecommandlockouts
\usepackage{cite}
\usepackage{amsmath,amssymb,amsfonts}
\usepackage{algorithmic}
\usepackage{graphicx}
\usepackage{textcomp}
\def\BibTeX{{\rm B\kern-.05em{\sc i\kern-.025em b}\kern-.08em
    T\kern-.1667em\lower.7ex\hbox{E}\kern-.125emX}}

\usepackage{soul}
\usepackage{booktabs} 
\usepackage[table]{xcolor}
\definecolor{grey}{rgb}{0.25, 0.25, 0.25}
\newcommand\crule[3][black]{\textcolor{#1}{\rule{#2}{#3}}}
\usepackage{amssymb}
\usepackage{wrapfig}
\usepackage[shortlabels]{enumitem}
\usepackage{tcolorbox}
\usepackage{balance}

\begin{document}

\title{Profiling US Restaurants from Billions of Payment Card Transactions \thanks{\textsuperscript{*}During this work, both authors were employed at Visa Research.}}

\author{
\IEEEauthorblockN{Himel Dev}
\IEEEauthorblockA{
\textit{University of Illinois (UIUC)}\\
Champaign, IL, USA\\
hdev3@illinois.edu}
\and
\IEEEauthorblockN{Hossein Hamooni}
\IEEEauthorblockA{\textit{Facebook Inc.} \\
Menlo Park, CA, USA\\
hamooni@fb.com}
}

\maketitle

\begin{abstract}
A payment card (such as debit or credit) is one of the most convenient payment methods for purchasing goods and services. Hundreds of millions of card transactions take place across the globe every day, generating a massive volume of transaction data. The data render a holistic view of cardholder-merchant interactions, containing insights that can benefit various applications, such as payment fraud detection and merchant recommendation. However, utilizing these insights often requires additional information about merchants missing from the data owner's (i.e., payment company's) perspective. For example, payment companies do not know the exact type of product a merchant serves. Collecting merchant attributes from external sources for commercial purposes can be expensive. Motivated by this limitation, we aim to infer latent merchant attributes from transaction data. As proof of concept, we concentrate on restaurants and infer the cuisine types of restaurants from transactions. To this end, we present a framework for inferring the cuisine types of restaurants from transaction data. Our proposed framework consists of three steps. In the first step, we generate cuisine labels for a limited number of restaurants via weak supervision. In the second step, we extract a wide variety of statistical features and neural embeddings from the restaurant transactions. In the third step, we use deep neural networks (DNNs) to infer the remaining restaurants' cuisine types. The proposed framework achieved a 76.2\% accuracy in classifying the US restaurants. To the best of our knowledge, this is the first framework to infer the cuisine types of restaurants by analyzing transaction data as the \textit{only} source.
\end{abstract}


\section{Introduction}
\label{sec:introduction}

Payment cards (primarily debit and credit), with billions of active users, are among the most popular payment methods across the globe. In the US alone, millions of people regularly use debit and credit cards to pay for goods and services. Payment processing companies handle these transactions and record their attributes (e.g., cardholder id, merchant name, and authorized amount), generating a massive volume of transaction data. Transaction data render a holistic view of cardholder-merchant interactions, containing insights into the behavior of cardholders and the business of merchants. The insights can benefit a variety of applications, such as payment fraud detection and merchant recommendation.

While transaction data accommodate valuable insights, utilizing these insights often require auxiliary information about merchants that are not present in raw data. For instance, in payment fraud detection, it is critical to know merchants' categories---the type of goods or services they provide---to classify and track purchases. Recommendation services also require characterizing merchant attributes, such as business category and sub-categories (the type of goods or services), market positioning (high-end vs. low-end), and business hours (when to visit). Broadly, incorporating merchant attributes can boost the performance of machine learning models that empower today's recommendation and anomaly detection services.

However, acquiring merchant attributes is challenging, especially for commercial purposes. Even payment processing companies that possess the most detailed view of transactions do not have access to all required attributes. One possible way to acquire the missing attributes is to collect them from crowd-sourcing companies (e.g., Yelp, Foursquare), which generate merchant profiles from customer reviews. However, such data acquisition has several limitations. First, it is costly to acquire merchant information for commercial purposes. Second, debit and credit cards are commonly used in many countries where collecting merchant information through crowd-sourcing may not be possible. For example, Yelp currently operates in 32 countries, whereas Visa cards are accepted in more than 200 countries and territories. Third, the collected information may become outdated, as new merchants appear or old merchants go out of business. These limitations motivate us to generate merchant profiles by analyzing transactions. Unlike user reviews, transaction data do not contain a large corpus of texts or images to extract merchant attributes. Consequently, generating merchant profiles from transaction data requires inferring attributes through other means.

\textbf{Present Work.} The present work contributes to our broad goal of inferring latent merchant attributes from transaction data. Though raw transactions do not contain much information about merchants, the transaction data as a whole contains rich signals, which have the potential to reveal a multitude of merchant attributes. As \textit{proof of concept}, we concentrate on restaurants and infer the cuisine types of restaurants from transactions. Restaurant is perhaps the most popular merchant category, with which most cardholders have regular transactions. However, while processing any restaurant transaction, a payment company does not know what type of cuisine the restaurant serves. This information can be useful for providing restaurant recommendations and improving fraud detection. 

\textbf{Restaurant Recommendation.} Cuisine types play a critical role in restaurant recommendation. For example, consider a transaction-based restaurant recommendation system that provides recommendations without being aware of restaurants' cuisine types. A standard solution to develop such a recommendation system is to employ user-based collaborative filtering, which generates recommendations based on the similarity of transactions among cardholders. While practical, this solution comes with considerable risk---imagine recommending steakhouses to vegetarians who visited many Mediterranean restaurants, as Mediterranean cuisine is popular among both vegetarians and meat-lovers. Incorporating cuisine types of restaurants into the recommendation algorithm can prevent such misreckoning. Cusine types can also be useful for addressing the cold-start problem in restaurant recommendation.

\textbf{Fraud Detection.} Cuisine types of restaurants also have the potential to inform fraud detection systems. For example, suppose a fraud detection system has access to the cuisine types of restaurants. In that case, it can use this information in conjunction with transaction data to generate a cuisine signature for each cardholder. This signature captures how likely the cardholder is to visit different types of restaurants. Then, while processing a restaurant transaction, the system can examine if the restaurant aligns with the cardholder's cuisine signature, among other signals, to detect if the transaction is a fraudulent one.

This paper presents a framework to infer the cuisine types of restaurants from billion-scale payment card transactions. \ul{Please note that the intended use of the framework includes commercial purposes. For this reason, we can not use external sources (e.g., Google, Yelp) for data acquisition.} The proposed framework has three steps: 1) weakly-supervised label generation, 2) statistical feature and neural embedding extraction, and 3) deep neural network based classification. In the first step, we programmatically generate cuisine labels for restaurants via weak supervision. We first compile a cuisine taxonomy containing the major cuisine types and a list of keywords for each cuisine. We then use the keywords as common patterns to label restaurants based on their names. For example, if the keyword ``Peking'' appears in a restaurant's name, we recognize it as a Chinese restaurant. Through bootstrapped label expansion, our weakly-supervised approach can generate cuisine labels for 35\% of US restaurants. We infer the remaining US restaurants' cuisine labels by employing large-scale feature engineering followed by deep neural network based classification. In the second step, we extract a wide variety of statistical features and neural embeddings for half a million US restaurants, from four billion restaurant transactions. Our statistical features encompass a wide range of business information about restaurants, such as location, business hours, pricing information, tipping culture, serving capacity, expected party size, customer visitation patterns, and customer loyalty. We also model transaction data as a bipartite graph between customers and restaurants and then employ state-of-the-art neural language models to generate two sets of restaurant embeddings. The embeddings capture the semantic similarity between restaurants at two distinct levels: the semantic similarity between any two restaurants that an individual customer visits (micro-level), the semantic similarity between two restaurants based on their entire customer base (macro-level). We further apply neural language models on restaurant names (excluding the words used in labeling) to generate another set of restaurant embeddings. Overall, we generate three sets of restaurant embeddings to represent the latent characteristics of restaurants. In the third step, we design several deep neural networks (DNNs) to predict remaining restaurants' cuisine types based on the statistical features and embeddings generated in the second step. Our best performing network is a deep feedforward network with residual connections. The network achieved a 76.2\% overall accuracy in classifying US restaurants, where per class accuracy is high for all classes, including minority cuisine types such as Mediterranean cuisine. To the best of our knowledge, this is the first attempt to profile restaurants in the mostly untapped space of payment card transactions. 
\section{Problem Setting}
\label{sec:problem}

This work aims to develop a framework for inferring the cuisine types of restaurants from debit and credit card transactions. Cuisine type information is essential to payment processing companies for providing user services, notably, restaurant recommendation and payment fraud detection. 

\textbf{Feature Engineering.} A secondary goal of this work is to extract general-purpose features that can benefit a variety of tasks beyond cuisine-based restaurant classification, such as business profiling and customer segmentation. Preferably, we want to extract interpretable features that can provide insights into the restaurant business.

\textbf{Data Restriction.} The intended use of this framework includes commercial purposes. Therefore, we can not use external sources (e.g., Google, Yelp) for label generation, feature engineering, and framework validation. For example, without explicit consent, Yelp API and content usage are restricted to non-commercial use~\cite{yelp2020}. Acquiring merchant information for commercial purposes can be costly. 

In remaining sections we present a framework that meets the requirements above and achieves a high accuracy in inferring the cuisine types of US restaurants.
\section{Transaction Dataset}
\label{sec:data}

Our dataset contains four billion debit and credit card transactions in more than half a million US restaurants within three months. This dataset is a large subset of in-person restaurant transactions in the US (customer made payment at a restaurant using a \textsc{Visa} card). Each transaction in our dataset is an interaction between a cardholder and a merchant. Therefore, all attributes of each transaction come from the cardholder, the merchant, or their interaction. Specifically, our dataset contains the following information for each transaction.

\begin{enumerate}[(a)]
\item \textbf{Merchant Name:} the legal name of the merchant.
\item \textbf{Merchant Location:} the ZIP code of the merchant.
\item \textbf{Timestamp:} the exact date and time of the transaction.
\item \textbf{Cardholder Id:} an anonymous id to group all transactions of the cardholder.
\item \textbf{Authorized Amount:} the amount (in USD) that the merchant initially charges.
\item \textbf{Settlement Amount:} the amount (in USD) that finally gets posted on the cardholder's bank statement. For restaurant transactions, the difference between the authorized amount and the settlement amount is the tip.
\end{enumerate}
\section{Cuisine Inference Framework}
\label{sec:method}

\begin{figure}[htb]
\centering
\includegraphics[scale=0.42]{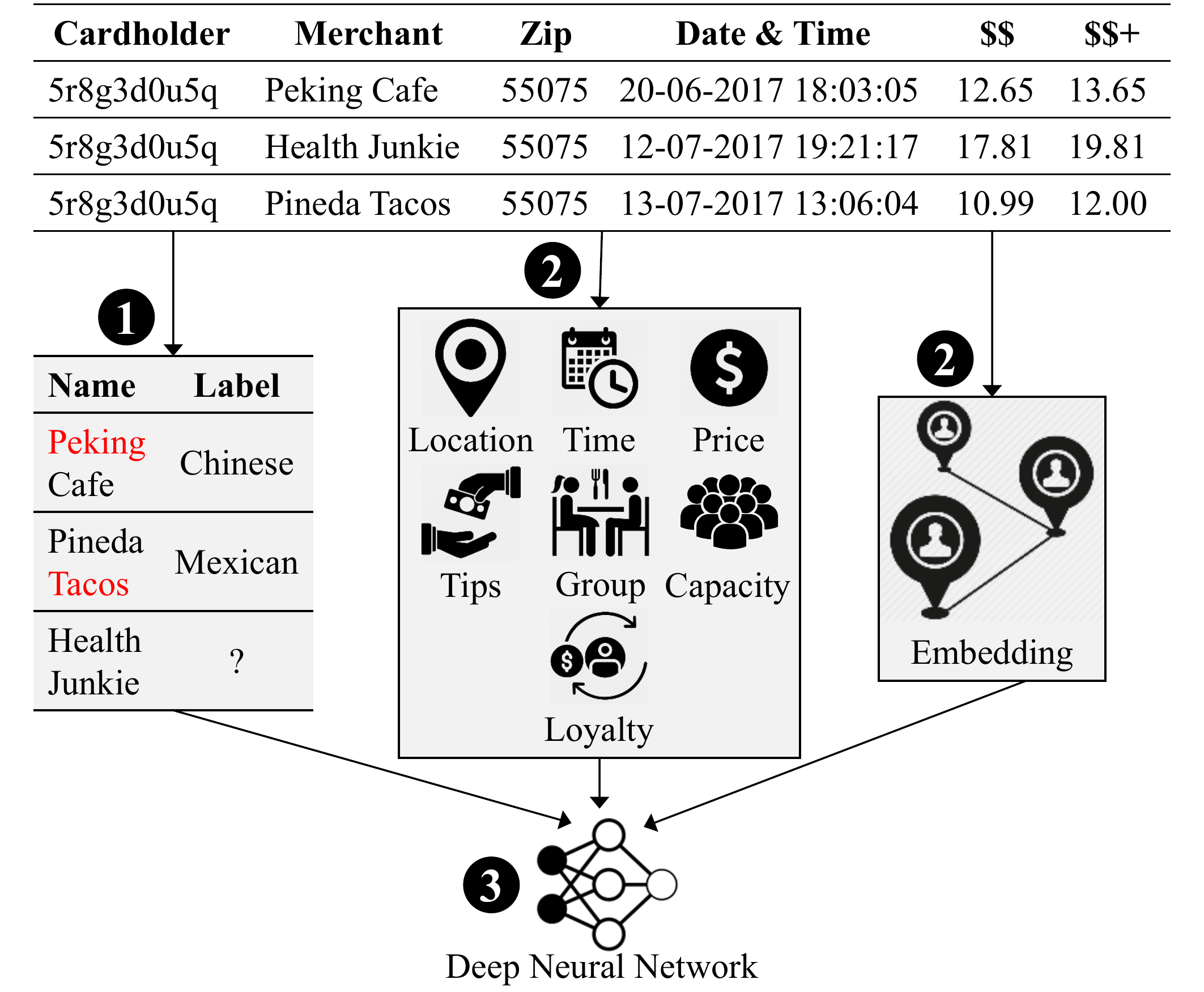}
\caption{Our proposed framework for inferring the cuisine types of restaurants. The framework has three steps: 1) weakly-supervised label generation, 2) statistical feature and neural embedding extraction, 3) deep neural network based classification.}
\label{fig:system}
\end{figure}

In this section, we describe our proposed framework for inferring the cuisine types of restaurants from debit and credit card transactions. Figure~\ref{fig:system} presents an overview of the framework. The framework has three steps: 1) weakly-supervised label generation, 2) statistical feature and neural embedding extraction, 3) deep neural network based classification. We perform these three steps in a \textit{closed-loop} to classify the restaurants in our transaction dataset. 

\subsection{Weakly-Supervised Label Generation}
In the first step, we programmatically generate cuisine labels for restaurants via weak supervision. We compile a cuisine taxonomy along with keywords, extract new keywords from restaurant names via bootstrapping, and employ topic modeling to augment the keyword-based approach. 

\textbf{Cuisine Taxonomy Compilation.} We compile a cuisine taxonomy based on the literature on food~\cite{Sajadmanesh2017WWW, Ahn2011Nature} and the US restaurant industry~\cite{Lee2014Food, Park2017Food, Somashekhar2018Migration}. Table~\ref{tab:cuisine} presents our cuisine taxonomy. Our taxonomy contains the ten most popular cuisine types in the US: Latin American, European, Middle Eastern \& Mediterranean \& African (MMA), South Asian, South East Asian, East Asian, Grill \& Steak, Fastfood, Bar, Dessert. Each of these major cuisine types (e.g., European) accommodates more fine-grained types (e.g., French). Though our taxonomy does not contain all cuisine types (e.g., Circassian), it has a high coverage for US restaurants. We compile a set of \emph{seed} words for each cuisine type in our taxonomy. We use these words as common patterns to generate cuisine labels for restaurant names. For example, if the keyword ``Peking'' appears in a restaurant's name, we recognize it as a Chinese restaurant. Currently, we have a list of 225 seed words that represent the ten major cuisine types. 


\begin{table}[thb]
	\caption{Manually crafted cuisine taxonomy for restaurant classification.}
    \label{tab:cuisine}
    \small
	\begin{center}    
	\begin{tabular}{ll}
    \toprule \textbf{Cuisine Type} & \textbf{Subcategories}\\ \midrule
    Latin American & Mexican, Cuban, Brazilian, Colombian\\ \midrule
	European & French, Italian, German, Polish, Irish\\ \midrule
    Mediterranean & Greek, Turkish\\ 
    Middle Eastern & Saudi Arabian, Lebanese, Persian, Afghan\\ 
    African & Moroccan, Ethiopian, Eritrean\\ \midrule
    South Asian & Indian, Pakistani, Nepalese, Bangladeshi\\ \midrule
    South East Asian & Thai, Vietnamese, Indonesian, Malaysian\\ \midrule
    East Asian & Chinese, Japanese, Korean, Mongolian\\ \midrule
    Grill and Steak & Grill, Steakhouse\\ \midrule
    Fastfood & Sandwich, Burger, Pizza\\ \midrule
    Bar & Bar, Pub, Tavern, Inn\\ \midrule
	Dessert & Ice Cream, Cafe, Bakery, Juice\\ \bottomrule
	\end{tabular}
    \end{center}
\end{table}

\textbf{Bootstrapped Label Expansion.} To better utilize the cuisine information obtainable from restaurant names, we apply pattern-based bootstrapping~\cite{Gupta2014CONNL}. The key idea of our approach is to extract new words (beyond seed words) from restaurant names, which can serve as highly accurate patterns to increase the coverage of labeled restaurants. We first determine the cuisine label distribution for each new word based on the labeled restaurants. We then select the words that meet three \textit{minimum} criteria---i) \textit{frequency:} the word needs to appear in $\theta_f$ fraction of restaurant names; ii) \textit{precision:} if we use the word and its majority label as a labeling rule, the rule needs to be true for $\theta_p$ fraction of labeled restaurants; iii) \textit{significance:} the ratio of labeled and unlabeled restaurants that contain the word should be $\theta_s$. We identify 198 words that satisfy the criteria mentioned above and use these bootstrapped words as secondary patterns to label more restaurants. Altogether, using the seed and bootstrapped keywords, we label 35\% of restaurants in our dataset. 

\begin{table*}[thb]
	\caption{The description of statistical features and neural embeddings extracted from restaurant transactions.}
    \label{tab:features}
    \small
	\begin{center}    
	\begin{tabular}{ll}
	\toprule \textbf{Feature Type} & \textbf{Description}\\ \midrule
	Pricing & The deciles of authorized amount in transactions\\
    Tipping culture & The deciles of (settlement amount - authorized amount) in transactions\\
    Serving capacity & The deciles of hourly transaction count\\
    Party size & The proportion of transactions for different party size\\
    Party pricing & The average authorized amount for different party size\\
    Temporal pattern I & The distribution of number of transactions over days of the week\\
    Temporal pattern II & The distribution of number of transactions over the hours of weekdays\\
    Temporal pattern III & The distribution of number of transactions over the hours of weekends\\
	Customer revisitation & The deciles of the number of revisits by the customers\\
    Customer loyalty &  The deciles of the number of restaurants visited by the customers\\
    Location & The digits of restaurant zipcode and corresponding location granularity\\
    Micro embedding & The restaurant embedding generated by applying word2vec on user documents\\
    Macro embedding & The restaurant embedding generated by applying doc2vec on restaurant documents\\
    Name embedding & The embedding based on restaurant name\\ \bottomrule
	\end{tabular}
    \end{center}
\end{table*}


\textbf{Topic Modeling for Weak Supervision.} To augment the keyword-based approach of weak supervision, we apply a variety of topic models on restaurant names, treating each restaurant name as a document. Conventional topic models such as LDA~\cite{blei2003latent} do not work well with restaurant names, resulting in incoherent topics, as measured via UMass coherence score~\cite{mimno2011optimizing}. We observe three critical challenges in applying topic models--- i) \textit{monolith:} many restaurant names consist of a single word; ii) \textit{sparsity:} sparse word co-occurrence patterns in restaurant names; iii) \textit{long-tail:} long-tail distribution of words appearing in restaurant names.  To address the monolith issue, we adopt the concept of \textit{sprinkling}~\cite{chakraborti2007supervised}---adding artificial words (known cuisine types) to boost the association in the topic model. To address the sparsity issue, we adopt the bi-term topic model (BTM)~\cite{Yan2013WWW} that learns topics by directly modeling the generation of word co-occurrence patterns known as biterms. BTM has two primary advantages over conventional topic models: i) BTM explicitly models word co-occurrence patterns, instead of word documents, which improves the topic learning process; ii) BTM uses the aggregated patterns in the whole corpus for learning topics, which solves the problem of sparse patterns at document-level. To address the long-tail issue, we perform \textit{stratified sampling} on biterms before passing them to BTM. Using BTM as our topic model, we vary the topic model parameter $k$ (the number of topics) from 5 to 50, while assigning the value $\frac{50}{k}$ to parameter $\alpha$ (document-topic density) and 0.1 to parameter $\beta$ (topic-word density), as suggested by Griffiths et al.~\cite{griffiths2004finding}. We then compute the UMass coherence score~\cite{mimno2011optimizing} for each set of topics. The resultant topics (cuisine classes) are coherent (high UMass score) and consistent with the cuisine classes generated by our keyword-based approach. 

\subsection{Statistical Feature and Embedding Extraction}
In the second step, we perform extensive feature engineering for large-scale transaction data. Table~\ref{tab:features} presents the description of the features that we extract from restaurant transactions. Broadly, our features fall into two categories: statistical features that capture business information about restaurants, and embeddings that capture restaurant's latent characteristics based on customer-restaurant interactions.

\textbf{Statistical Feature Extraction.} We extract a wide variety of statistical features from transactions based on the literature on the restaurant industry~\cite{Fine1990Temporal, Qu1997Party, Muller1999Capacity, Clark1999Loyalty, Kimes2004Party, Kimes2004Party2, Ryu2011Repeat, Barber2011Repeat, Schlosser2012Speed, She2012Location, Yim2014Hospitality, Ray2016Ethnic, Chen2017Tipping}. These features encompass a wide range of business information about restaurants, such as location, business hours, pricing information, tipping culture, serving capacity, expected party size, customer visitation patterns, and customer loyalty. The features provide insights into the restaurant business and benefit various tasks beyond restaurant classification, such as business profiling and customer segmentation. A key idea of our statistical feature engineering approach is to use distributions as features, enabling the framework to differentiate among cuisine types in terms of tail behavior.

\begin{table*}[hbt]
	\caption{Difference among cuisine types in terms of statistical features. Percentile based color coding: \crule[grey!5]{0.25cm}{0.25cm} low, \crule[grey!15]{0.25cm}{0.25cm} medium, \crule[grey!25]{0.25cm}{0.25cm} high}
    \label{tab:insights}
    \small
	\begin{center}    
	\begin{tabular}{m{3cm}m{1.3cm}m{1.3cm}m{1.3cm}m{1.3cm}m{1.3cm}m{1.3cm}m{1.3cm}m{1.3cm}}
    \toprule
    \vspace{1.3cm}\textbf{1. Cuisine Type} & \rotatebox[origin=c]{90}{\parbox{2.9cm}{\textbf{2. Price\\(Median)}}} & \rotatebox[origin=c]{90}{\parbox{2.9cm}{\textbf{3. Tips\\(Median)}}} & \rotatebox[origin=c]{90}{\parbox{2.9cm}{\textbf{4. Serving Capacity\\(Mean)}}} & \rotatebox[origin=c]{90}{\parbox{2.9cm}{\textbf{5. Customer Revisit\\(90th Percentile)}}} & \rotatebox[origin=c]{90}{\parbox{2.9cm}{\textbf{6. Restaurant Tally\\(Median)}}} & \rotatebox[origin=c]{90}{\parbox{2.9cm}{\textbf{7. Expense/Person\\(Mean)}}} & \rotatebox[origin=c]{90}{\parbox{2.9cm}{\textbf{8. Single Diner\\(Percentage)}}} & \rotatebox[origin=c]{90}{\parbox{2.9cm}{\textbf{9. Weekend Tranx.\\(Percentage)}}}\\\midrule
    Dessert & \cellcolor{grey!5}8.13 & \cellcolor{grey!5}0 & \cellcolor{grey!15}24 & \cellcolor{grey!25}32 & \cellcolor{grey!25}19 & \cellcolor{grey!5}5 & \cellcolor{grey!25}46 & \cellcolor{grey!5}48\\
    European & \cellcolor{grey!25}28.89 & \cellcolor{grey!25}4 & \cellcolor{grey!15}23 & \cellcolor{grey!5}17 & \cellcolor{grey!5}17 & \cellcolor{grey!25}17 & \cellcolor{grey!25}45 & \cellcolor{grey!25}53\\
    Fastfood & \cellcolor{grey!5}16.45 & \cellcolor{grey!5}0 & \cellcolor{grey!5}19 & \cellcolor{grey!5}19 & \cellcolor{grey!5}17 & \cellcolor{grey!5}9 & \cellcolor{grey!15}42 & \cellcolor{grey!15}50 \\
    Bar & \cellcolor{grey!15}22.53 & \cellcolor{grey!25}4 & \cellcolor{grey!25}27 & \cellcolor{grey!25}30 & \cellcolor{grey!15}18 & \cellcolor{grey!15}13 & \cellcolor{grey!25}45 & \cellcolor{grey!25}54 \\ 
    Grill \& Steak & \cellcolor{grey!15}21.91 & \cellcolor{grey!15}3 & \cellcolor{grey!25}26 & \cellcolor{grey!15}24 & \cellcolor{grey!15}18 & \cellcolor{grey!15}12 & \cellcolor{grey!15}44 & \cellcolor{grey!15}51 \\ 
    Latin American & \cellcolor{grey!5}16.69 & \cellcolor{grey!5}0 & \cellcolor{grey!15}23 & \cellcolor{grey!15}22 & \cellcolor{grey!15}18 & \cellcolor{grey!15}10 & \cellcolor{grey!15}43 & \cellcolor{grey!5}48 \\ 
    MMA & \cellcolor{grey!5}16.57 & \cellcolor{grey!5}0 & \cellcolor{grey!5}20 & \cellcolor{grey!5}18 & \cellcolor{grey!25}19 & \cellcolor{grey!15}11 & \cellcolor{grey!25}46 & \cellcolor{grey!5}46 \\ 
    South Asian & \cellcolor{grey!15}23.87 & \cellcolor{grey!15}2 & \cellcolor{grey!5}18 & \cellcolor{grey!5}16 & \cellcolor{grey!15}18 & \cellcolor{grey!15}13 & \cellcolor{grey!15}40 & \cellcolor{grey!25}52 \\ 
    South East Asian & \cellcolor{grey!15}21.88 & \cellcolor{grey!15}2 & \cellcolor{grey!5}17 & \cellcolor{grey!5}16 & \cellcolor{grey!25}19 & \cellcolor{grey!15}11 & \cellcolor{grey!5}35 & \cellcolor{grey!5}47 \\ 
    East Asian & \cellcolor{grey!15}19.40 & \cellcolor{grey!5}0 & \cellcolor{grey!5}17 & \cellcolor{grey!5}15 & \cellcolor{grey!5}17 & \cellcolor{grey!15}10 & \cellcolor{grey!5}39 & \cellcolor{grey!5}48 \\ 
    \bottomrule
	\end{tabular}
    \end{center}
\end{table*}

\textbf{Pricing:} Price is one of the crucial features for characterizing a restaurant, which bears implications for cuisine type~\cite{Yim2014Hospitality}. Roy~\cite{Ray2016Ethnic} reported the price inequality among different cuisine types for restaurants located in New York City (NYC), e.g., people in NYC pay more for European cuisines compared to East Asian and South Asian cuisines. To capture the idiosyncrasies of pricing, we extract the deciles of the authorized amount for each restaurant. The 2\textsuperscript{nd} column in Table~\ref{tab:insights} shows how the median price of food varies across cuisine types, for all US restaurants (median of median). The median price is: a) high for European cuisines; b) medium for Bar, Grill \& Steak, and Asian cuisines; c) low for Dessert, Fastfood, Latin American, and MMA cuisines. 

\textbf{Tipping Culture:} Tipping norms vary across restaurant types, e.g., fine dining restaurants have different norms compared to fast food restaurants. Chen et al. ~\cite{Chen2017Tipping} examined how the bias towards different types of cuisines affect gratuities. The study confirms the effect of cuisine types on tipping behavior. To capture customers' tipping behavior for a restaurant, we extract the deciles of the difference between the settlement amount and the authorized amount. The 3\textsuperscript{rd} column in Table~\ref{tab:insights} shows how the median of tips varies across cuisine types. The median tip is: a) high for Bar and European cuisines; b) medium for Grill \& Steak, South Asian, and South East Asian cuisines; c) low for all other cuisines. 

\textbf{Serving Capacity:} Serving capacity is an important variable for studying a restaurant business~\cite{Muller1999Capacity}, e.g., fast food restaurants typically have higher serving capacity compared to the fine dining restaurants~\cite{Schlosser2012Speed}. To capture the serving capacity of a restaurant, we extract the deciles of the number of transactions per hour. The 4\textsuperscript{th} column in Table~\ref{tab:insights} shows how the average number of servings varies across cuisine types, for all US restaurants (median of average). Average serving capacity (number of customers per hour) is: a) high for Bar, and Grill \& Steak; b) medium for Dessert, European, and Latin American cuisines; c) low for all other cuisines.

\textbf{Party Size:} Party size composition is an insightful feature for studying the customers of a restaurant~\cite{Kimes2004Party, Kimes2004Party2}. Qu et al.~\cite{Qu1997Party} showed that at Chinese restaurants in Indiana, the majority (57\%) of customers dined with a party size of two. Estimating party size from transaction data is challenging. A bill for a dining party could be paid by one of the members or shared among the members. If one of the members pays the entire bill, there would be no explicit information about the party size. In contrast, if the members share the bill, there would be no association between their transactions. We overcome the first challenge by estimating party size from the authorized amount. We can not address the second challenge due to the weak association between transactions. To estimate party size, we apply a separate Gaussian mixture model (GMM) on each restaurant's authorized amounts. Our Gaussian mixture model is a weighted sum of $K$ component Gaussian densities as given by the equation,

$$p(x|\theta) = \sum_{i=1}^{K} \phi_i \mathcal{N}(\mu_i, \sigma_i),$$

where $x$ is the vector of authorized amounts, $\phi_i$ is the mixture weight (prior probability) of $i$-th component, and $\mathcal{N}(\mu_i, \sigma_i)$ is the Gaussian density of $i$-th component. We determine the \textit{optimal} number of mixture components for GMM by using the Akaike information criterion (AIC). We use the proportion of transactions for different party sizes, and the corresponding average authorized amounts as party composition features. The 7\textsuperscript{th} column in Table~\ref{tab:insights} shows how the average expenditure per person varies across cuisine types. Consistent with pricing, the average expenditure per person is highest for European cuisine and lowest for dessert. The 8\textsuperscript{th} column in Table~\ref{tab:insights} shows how the percentage of single diner varies across cuisine types. Notice that the single diner percentage is lower for the Asian cuisines compared to other cuisine types. This finding is consistent with the norm that Asian restaurants are popular for group eating.

\textbf{Temporal Patterns:} Temporal patterns such as business hours and busy times are critical indicators for understanding a restaurant business~\cite{Fine1990Temporal}, e.g., bars typically operate during night time, whereas brunch spots operate during day time. To capture the temporal patterns of a restaurant's operation, we extract the distribution of transactions over days-of-week and hours-of-day. We further divide the hours-of-day into hours-of-weekday and hours-of-weekend. The 9\textsuperscript{th} column in Table~\ref{tab:insights} shows how the percentage of transactions on weekends varies across cuisine types, for all US restaurants. Notice that Bars and European restaurants attract more of their customers during weekends.

\textbf{Customer Revisitation:} Researchers have acknowledged the importance of understanding the differences between first-time customers and repeat customers in restaurants~\cite{Ryu2011Repeat}. Barber et al.~\cite{Barber2011Repeat} showed a strong correlation between the food type and repeat visitation of customers. To capture the customer revisitation tendency in a restaurant, we extract the deciles of the number of revisits (repeat transactions) by the customers. The 5\textsuperscript{th} column in Table~\ref{tab:insights} shows how the number of revisits for top customers (on and above 90th percentile) varies across cuisine types. Notice that frequent customers have a strong revisit tendency to bars and dessert places.

\textbf{Customer Loyalty:} Customer loyalty can be useful for studying a restaurant business~\cite{Clark1999Loyalty}. This feature is especially helpful for understanding the customers and their endemicity. To capture the loyalty of customers of a restaurant, we extract the deciles of the number of restaurants visited by the customers.  The 6\textsuperscript{th} column in Table~\ref{tab:insights} shows how the median number of restaurants visited by customers varies across cuisine types. Notice that there is not much difference among the cuisine types in terms of customer base loyalty.

\textbf{Location:} Location has been a critical factor for characterizing a restaurant~\cite{She2012Location}. As a restaurant business is often established based on potential customers' preference in the locality, location plays a crucial role in restaurant profiling. We extract the digits of restaurant zip code and corresponding location granularity to capture a restaurant's locality.

\textbf{Restaurant Embedding Generation.} Restaurants in our dataset can be represented in a lower-dimensional space based on the inherent structure imposed by their characteristics, such as cuisine type, price range, and location. Cardholders can also be represented in a lower-dimensional space based on their traits, such as cuisine affinity, price sensitivity, and location. The transactions in our dataset are interactions between samples from these two spaces, where certain interactions are plausible while others are unlikely to happen. To capture the interactions between cardholders and restaurants, we model transaction data as a bipartite graph. We then propose two hypotheses that capture the plausibility of the interactions in this graph at the micro and macro level.  


\begin{tcolorbox}
\ul{\textbf{Hypothesis I:}} \textit{The plausibility of interaction between a customer and a restaurant is a function of the customer's preferences and the restaurant's attributes.} 
\end{tcolorbox}
\begin{tcolorbox}

\ul{\textbf{Hypothesis II:}} \textit{The plausibility of interaction between a group of customers and a restaurant is a function of the collective taste and the restaurant's attributes.} 
\end{tcolorbox}

The subtle distinction between the two hypotheses lies in their application: the first hypothesis applies to individuals (micro), and the second hypothesis applies to groups (macro). The first hypothesis implies that the compatibility between a customer's preferences and a restaurant's attributes is a good predictor of whether the customer will visit the restaurant. For example, a vegetarian is likely to visit an Indian restaurant. The second hypothesis implies that the type of customers who visit a given restaurant is a good predictor of its attributes. For example, a restaurant is unlikely to be a steakhouse if many of its customers are vegetarian.

Based on the hypotheses mentioned above, we aim to represent the restaurants in our dataset in a low-dimensional space by applying state-of-the-art representation learning techniques. There is a variety of methods that can be used for learning lower-dimensional representations---notably neural language models, such as word2vec~\cite{Mikolov2013NIPS} and its extension doc2vec~\cite{Le2014ICML}. These models have been successfully adopted in a variety of domains, including product representation in eCommerce~\cite{Grbovic2015}. In this work, we utilize word2vec and doc2vec to create two distinct sets of restaurant embeddings based on our micro and macro hypotheses. Details about the parameters of the micro (word2vec) and macro (doc2vec) model can be found in the experimental evaluation (impact of hyperparameters).

\textbf{Micro Embedding:} We represent each customer as a document and each restaurant as a word. Accordingly, each transaction by a customer corresponds to a word (restaurant) in a document (the customer's ledger of restaurants). The words appear chronologically in the documents, as per the day and time of transactions. After creating the documents, we apply word2vec on these documents. As per our representation, words refer to restaurants, and consequently, the word embeddings generated by word2vec are restaurant embeddings. These embeddings capture our micro hypothesis: the context of a restaurant in a customer's ledger is defined by other restaurants visited by the customer within the same time window.

\textbf{Macro Embedding:} We represent each restaurant as a document and each customer as a word. Accordingly, each transaction in a restaurant corresponds to a word (customer) appearing in a document (the restaurant's ledger of customers). These words appear chronologically, as per the day and time of transactions. After creating the documents, we apply doc2vec on these documents. As per our representation, documents refer to restaurants, and consequently, the document embeddings returned by doc2vec are restaurant embeddings. These embeddings capture our macro hypothesis: the context of a customer in a restaurant is defined by other customers visiting the restaurant within the same time window.

\textbf{Name Embedding:} Apart from the two sets of embeddings learned from the customer-restaurant interactions; we generate another set of embeddings from restaurant names. Name embedding has been successfully utilized in other domains, such as social media \cite{ye2019secret}. In this paper, we generate name embeddings to utilize the remaining (non-labeling) words in restaurant names. To this end, we first remove the labeling words (used in step 1) from restaurant names. We then use pre-trained word embeddings (GloVe~\cite{pennington2014glove}) to generate name embeddings for restaurants. Specifically, to generate the name embedding of a restaurant, we combine the embeddings corresponding to its words via max pooling.

\subsection{DNN Based Classification}
\begin{figure}[htb]
\centering
\includegraphics[scale=0.325]{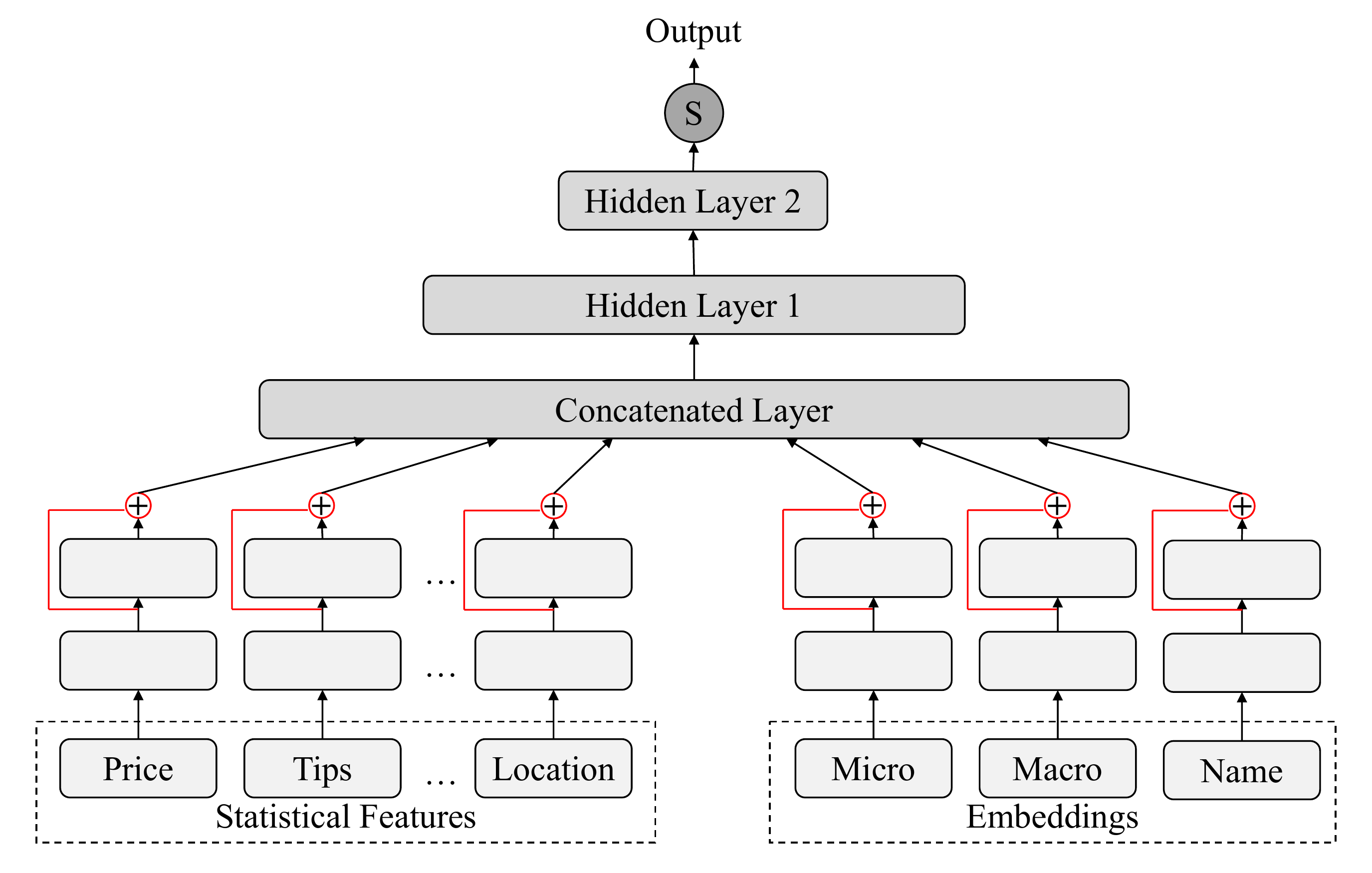}
\caption{Architecture of the best performing DNN model: a deep feedforward network with residual connections (marked in red).}
\label{fig:nn_architecture}
\end{figure}

In the third step, we classify the unlabeled restaurants in our dataset. To this end, we create several deep neural networks (DNNs) to complement our extracted features. Some of these networks mimic prominent DNN models deployed in large-scale industry setup, such as wide and deep~\cite{Cheng2016DLRS}, deep and cross~\cite{Wang2017ADKDD}. Here we only discuss the architecture of our best-performing DNN model: a deep feedforward network with residual connections. Figure~\ref{fig:nn_architecture} presents our best performing DNN model. The best performing DNN consists of several parallel paths, each corresponding to a different feature. First, we pass each of our features (vectors of real-numbers) through two hidden layers, where the layers have residual connections~\cite{He2016CVPR}. We then concatenate the resultant layers to create a single concatenated layer. Finally, we pass the concatenated layer through two more hidden layers and apply SoftMax activation at the output layer.

\section{Experimental Evaluation}
\label{sec:evaluation}

In this section, we provide a comprehensive evaluation of our cuisine inference framework. We conduct extensive experiments on a large-scale transaction dataset to validate the effectiveness of the framework. Specifically, we focus on answering the following research questions.

\begin{enumerate}
  \item How effective is our framework in inferring the cuisine types of restaurants?
  \item How useful are different feature types for classifying restaurants?
  \item What are the impacts of different hyper-parameters in overall performance?
\end{enumerate}

\subsection{Experimental Setup}

We first describe our experimental setup: evaluation dataset and DNN models.

\textbf{Evaluation Dataset.} Our transaction dataset contains $\sim$4,000,000,000 transactions in $\sim$650,000 restaurants. A sizable fraction of these restaurants are chains, e.g., McDonald's, Subway, and KFC. For the purpose of evaluation, we exclude the chain restaurants. The final evaluation dataset contains $\sim$500,000 restaurants. We generate cuisine labels for 35\% of these restaurants using our weakly-supervised method. We then split the labeled restaurants into training (80\%) and testing (20\%) sets. We used 5-fold cross-validation for tuning the hyperparameters. In addition to the weakly supervised restaurants, we manually label another 1000 restaurants to create a holdout set.

\textbf{DNN Models.} To demonstrate the effectiveness of our framework, we develop several DNN models as follows. 

\begin{enumerate}[(a)]
\item \textbf{Wide \& Deep:} This is a variant of the wide and deep network~\cite{Cheng2016DLRS} that captures the interaction between features. 

\item \textbf{Deep \& Cross:} This is a variant of the deep and cross network~\cite{Wang2017ADKDD} that applies feature crossing. 

\item \textbf{Shallow Feedforward:} This is a feedforward neural network with two hidden layers. We first pass each feature through a hidden layer, concatenate them, and finally pass them through another hidden layer. 

\item \textbf{Deep Feedforward:}  This is a feedforward neural network with four hidden layers. We first pass each feature through two hidden layers, then concatenate them, and finally pass them through two more hidden layers. 

\item \textbf{Deep Feedforward with Residual:} This is a deep feedforward network with residual connections~\cite{He2016CVPR}. We first pass each feature through two hidden layers (with residual connection between the two layers), concatenate them, and finally pass them through two more hidden layers.
\end{enumerate}

\subsection{Effectiveness of Our Framework (RQ1)}

We now discuss the effectiveness of our framework in inferring the cuisine types of restaurants.

\textbf{DNN Performance Comparison.} We compare the performance of our DNN models for inferring the cuisine types of restaurants. Table~\ref{tab:performance} reports the accuracy of our DNN models. These scores are based on the best configuration of hyperparameters (dropout, batch size, and learning rate) determined through a grid search. From the results, we make the following observations. 

\begin{enumerate}
    \item The Deep Feedforward network outperforms the Shallow Feedforward network as expected. The performance gap between the two is significant.
    \item Surprisingly, the Deep Feedforward network also outperforms the Wide \& Deep and Deep \& Cross networks. In fact, the Wide \& Deep network performs worse even compared to the Shallow Feedforward network.
    \item Adding residual connection between layers improves the performance of DNNs, as demonstrated by the Deep Feedforward with Residual network. 
    \item While the performance gain from using the best-performing architecture may seem small in a relative scale, the gain is significant in absolute terms: 1\% improvement implies correctly classifying another 5000 restaurants.
\end{enumerate}

\begin{table}[hbt]
	\caption{Comparison of accuracy among DNN models.}
    \label{tab:performance}
    \small
	\begin{center}    
	\begin{tabular}{lc}
    \toprule \textbf{Neural Architecture} & \textbf{Accuracy}\\ \midrule
    Wide \& Deep & 0.740 \\
    Deep \& Cross & 0.746 \\ 
    Shallow Feedforward &  0.743 \\
    Deep Feedforward & 0.756 \\
    Deep Feedforward with Residual & \textbf{0.762} \\ \bottomrule
	\end{tabular}
    \end{center}
\end{table}

\textbf{Pre-trained Embedding Baseline.} \textsc{Visa} previously generated a set of general-purpose merchant embeddings for all merchants (including restaurants). We used the pre-trained restaurant embeddings in conjunction with a deep feedforward network to develop a baseline for inferring the cuisine types of restaurants. This baseline achieved an accuracy of 0.51, implying that the proposed framework provides a 49\% improvement over the baseline.

\textbf{Non-Neural Baselines.} For completeness, we also compare the performance of our DNN models with several non-neural baselines. Our non-neural baselines include logistic regression and SVM with different kernel functions (linear, polynomial, rbf, and sigmoid). The best performing non-neural baseline is the multi-class SVM with a linear kernel, which achieves an accuracy of 0.351.

\textbf{Top-k Accuracy.} We investigate if further improvement in accuracy can be achieved by increasing the number of predictions. We compute top-k accuracy for our best-performing model, where correct prediction is assumed if the ground truth label appears in the top-k predictions. The top-2 and top-3 accuracy scores for our best-performing model are 0.879 and 0.921, respectively.

\begin{figure}[hbt]
\centering
\includegraphics[scale=0.6]{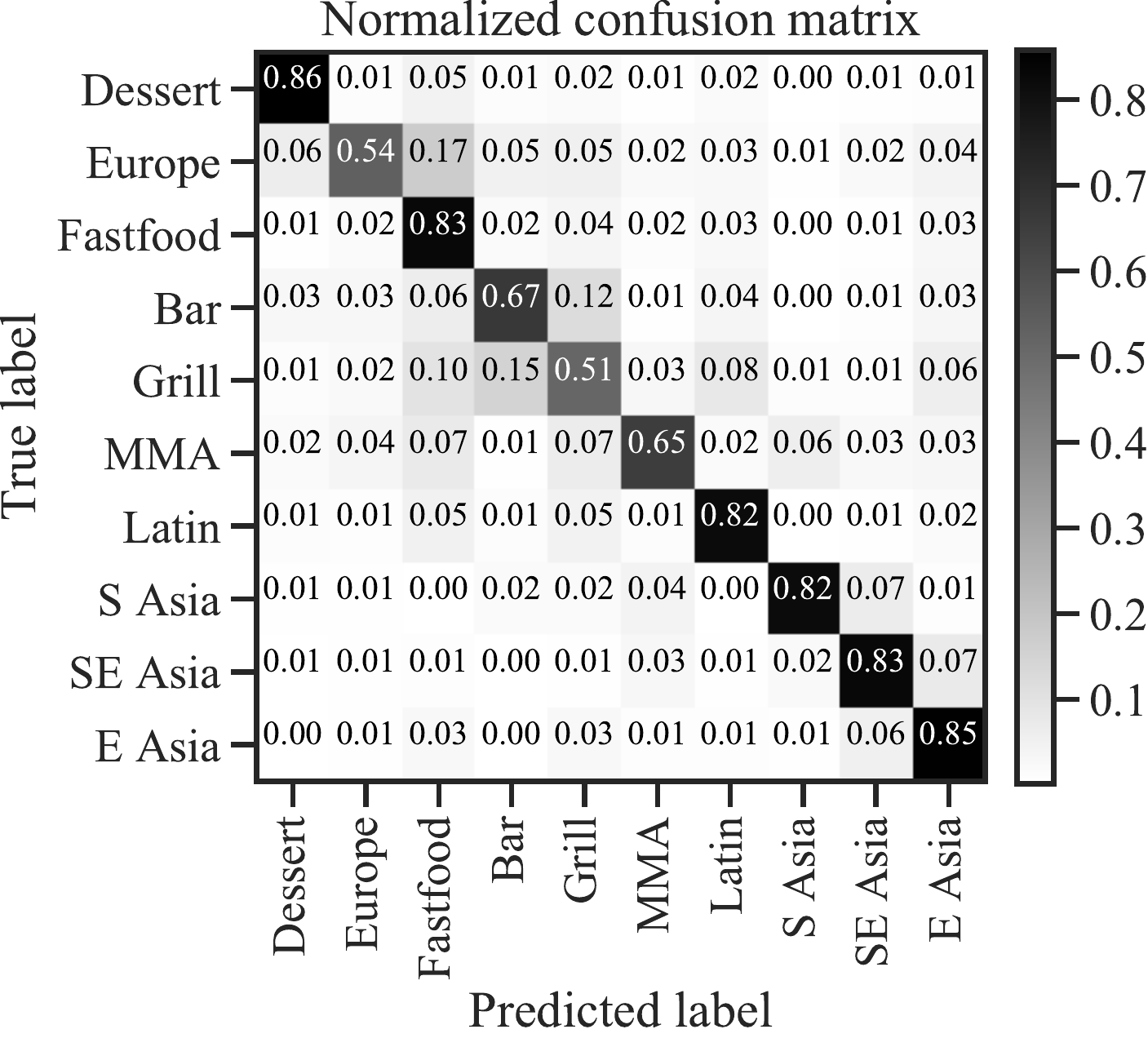}
\caption{Confusion matrix for the best performing model in the validation test. Notice that our framework leads to a \textit{balanced} confusion matrix, where per class accuracy is high for all classes.}
\label{fig:conf_mat}
\end{figure}

\textbf{Confusion Matrix.} As our dataset is highly imbalanced (class imbalance ratio up to 16:1), we study per class accuracy to better understand performance. Figure~\ref{fig:conf_mat} presents the normalized confusion matrix for the best performing model in the validation test. Notice that our framework leads to a \textit{balanced} confusion matrix, where per class accuracy is high for all classes, including minority classes such as South Asian. 

\textbf{Holdout Performance.} As we use the weakly-supervised labels as ground truth, the labels may introduce some bias. To investigate if such bias affects our final results, we use the manually labeled holdout set to review our framework's performance. We found that our framework's performance in the holdout set is comparable with the performance in the validation set.

\textbf{Model Stability.} As transaction data is time-varying, we investigate if our framework's predictions remain stable over time. To this end, we use another evaluation dataset (with six billion transactions) based on a different three months period. We generate predictions for unlabeled restaurants based on both datasets and compare the results. We observe that the predictions of our model remain stable over time.

\subsection{Importance of Features (RQ2)}

We perform an ablation study to understand the importance of different features in cuisine-based restaurant classification. Table~\ref{tab:ablation} reports the change in accuracy scores obtained by removing one feature at a time. We observe that the neural restaurant embeddings are the most useful features for cuisine type prediction. This finding implies that the embeddings successfully capture the latent characteristics of restaurants. Among the embeddings, the contribution of name embedding is particularly noteworthy. It shows that the remaining words (that have not been used for labeling) in restaurant names contain strong signals about cuisine type. The micro embedding that captures compatibility between a customer and a restaurant is another strong indicator, implying that many customers have a strong preference towards certain cuisine types. The macro embedding is also useful in cuisine type prediction, confirming the taste similarity among the customers who visit a given restaurant. Among the statistical features, temporal patterns play a crucial role, indicating that cuisine types impact the business hours and busy times. Other prominent statistical features are price, location, and customer visitation patterns. As evidenced by the decline in accuracy score, all of the proposed statistical features are also useful in cuisine type prediction.

\begin{table}[htb]
	\caption{Ablation study of features: change in accuracy after removing each feature.}
    \label{tab:ablation}
    \small
	\begin{center}    
	\begin{tabular}{lc}
    \toprule \textbf{Removed Feature} & \textbf{Accuracy (Change)}\\ \midrule
    Name embedding & 0.689 (-0.073)\\
    Micro embedding & 0.720 (-0.042)\\
    Macro embedding & 0.750 (-0.012)\\
    Transactions over hours-of-weekend & 0.750 (-0.012)\\
    Transactions over hours-of-weekday & 0.751 (-0.011)\\
    Transactions over days-of-week & 0.751 (-0.011)\\
    Price quantiles &  0.753 (-0.009)\\
    Zipcode based location hierarchy & 0.754 (-0.008)\\
    Tips quantiles & 0.755 (-0.007)\\
    Revisitation quantiles & 0.755 (-0.007)\\
    Average spending for party size & 0.756 (-0.006)\\
    Customer count quantiles & 0.757 (-0.005)\\
    Affinity quantiles & 0.757 (-0.005)\\
    Party size composition & 0.758 (-0.004)\\ \bottomrule
	\end{tabular}
    \end{center}
\end{table}

\subsection{Impact of Hyperparameters (RQ3)}

There are several sets of hyperparameters in our framework that may impact the overall performance, notably, the hyperparameters of the DNN model and the hyperparameters associated with the neural embeddings. To determine the impact of these hyperparameters, we perform a grid search over the space of hyperparameters. Table~\ref{tab:hyperparamters} shows the optimal configuration of hyperparameters (in bold). We found that two hyperparameters have a profound impact on performance: batch size (smaller batch size leads to significant improvement in performance) and the dimension of micro embedding (the dimension size significantly affects its contribution).

\begin{table}[htb]
	\caption{Optimal configuration of hyperparameters (in bold).}
    \label{tab:hyperparamters}
    \small
	\begin{center}    
	\begin{tabular}{ll}
    \toprule \textbf{DNN} & \\ \midrule
    Dropout &  \textbf{0.0}, 0.1, 0.2, 0.3\\
    Batch Size &  \textbf{128}, 256, 512, 1024\\
    Learning Rate &  0.001, 0.01, \textbf{0.1}, 1.0\\ \midrule
    \textbf{Micro Embedding} & \\ \midrule
    Dimension &  50, 100, \textbf{200}, 300\\
    Window Size &  5, 10, \textbf{20}, 30\\
    Negative Sampling &  5, 10, \textbf{20}, 30\\ \midrule
    \textbf{Macro Embedding} & \\ \midrule
    Dimension &  50, \textbf{100}, 200, 300\\
    Window Size &  50, 100, \textbf{200}, 300\\
    Negative Sampling &  5, 10, \textbf{20}, 30\\ \bottomrule
	\end{tabular}
    \end{center}
\end{table}

\section{Discussion}

We make the following observations from our results: 1) names are informative in identifying the cuisine types of restaurants; 2) different cuisine types considerably differ across the business dimensions (such as pricing, tipping culture, and serving capacity); 3) it is difficult to distinguish certain cuisine types (say South Asian and South East Asian) from one another. Now, we qualitatively analyze these results and present the implications of our findings.

\textbf{Cuisine Information in Restaurant Names.} Restaurant names often contain rich information about the underlying cuisine types. Recall that, with just 423 keywords (225 seed, 198 bootstrapped), we could label 35\% of US restaurants. Further, we validated the efficacy of the labeling through a holdout set. Altogether, the keywords provide high support and confidence in labeling restaurants. We also generated name embeddings form the remaining words in the restaurant names, which became the most informative feature in inferring the cuisine types. Overall, names have the potential to inform the latent characteristics of restaurants.

\textbf{Cuisine Types across Business Dimensions.} Our study reveals several insights about the US restaurant industry. Table~\ref{tab:insights} shows some of the insights derived from our statistical features. A careful inspection of these statistics presented reveals the inequalities across different ethnic cuisine types in the US. For example, across all US restaurants, the median tips for European cuisine are much higher compared to the Asian cuisines. These insights could open up new threads of research on the restaurant industry.

\textbf{Similarity of Cuisines.} We analyze the normalized confusion matrix to get an understanding of which classes are harder to separate. We observe that it's hard to separate the following classes: i) MMA from Fast food; ii) MMA from Grill \& Steakhouse; iii) South Asian from South East Asian; iv) South East Asian from East Asian; v) European from Fast food; vi) Bar from Grill \& Steakhouse; vii) Grill \& Steakhouse from Fast food; viii) Grill \& Steakhouse from Bar. Note that iii) and iv) are examples of kissing cuisines~\cite{Sajadmanesh2017WWW}: cuisines that are similar in terms of ingredients and flavors. In remaining cases, there are many restaurants that serve both types of cuisines. For example, many Mediterranean places serve gyros (a popular fast food), bars often include grills.
\section{Related Work}
\label{sec:related}

Our work draws from and improves upon, prior works on restaurant recommendation.

\textbf{Restaurant Profiling and Recommendation.} There is a recent body of work on restaurant profiling and recommendation~\cite{Lian2017WWW,Chu2017WWWJ,Zhang2016WWW,Bakhshi2014WWW,fu2014user}. These works have typically relied on user ratings and reviews for generating restaurant profiles and subsequent recommendations. Bakhshi et al.~\cite{Bakhshi2014WWW} studied the effects of restaurant attributes, local demographics, and local weather conditions at the date of visit on online customer reviews. Zhang et al.~\cite{Zhang2016WWW} designed a unified Collective Implicit Explicit Preference Model (CIEPM) to combine the implicit (check-in) and explicit (review) preference of customers for restaurant recommendation. Lian et al.~\cite{Lian2017WWW} used restaurant attributes, check-in data, and review text to perform a survival analysis of restaurants in China. We explore the problem of restaurant profiling in an untapped space---from debit and credit card transactions. Unlike user reviews, transaction data does not contain a large corpus of texts or images to infer attributes. Accordingly, we perform an extensive feature engineering for transaction data to infer restaurant attributes. 

\textbf{Studies on Restaurant Industry.} Our work draws from the rich literature on the restaurant industry. In addition to the works already mentioned, our study is motivated by the seminal work of Susan Auty~\cite{Auty1992SIJ}, and its followups~\cite{clark1998consumer}. The key finding of Auty's work is that food (cuisine) type is a primary factor in restaurant selection by consumers. Our proposed work builds upon this premise, where we identify cuisine types of US restaurants to provide meaningful restaurant recommendations
\section{Conclusion}
\label{sec:conclusion}

In this paper, we present a framework for inferring the cuisine types of restaurants from debit and credit card transactions, drawing from the literature on the restaurant industry, and applied machine learning. Our framework involves three steps---programmatically generating cuisine labels for restaurants, extracting statistical features and neural embeddings from transactions, and finally, deep neural network based classification. Our framework achieved a 76.2\% accuracy in classifying the US restaurants. Our findings have implications for feature engineering for large-scale transaction data, inferring latent merchant attributes, and studying the US restaurant industry. To the best of our knowledge, this is the first framework that performs cuisine inference by analyzing transaction data as the \emph{only} source.

\section*{Acknowledgment}
The authors would like to thank Mahashweta Das, Konik Kothari, Manoj Reddy, Aravind Sankar, and Hao Yang for their valuable feedback in this project.

\bibliographystyle{IEEEtran}
\balance
\bibliography{cuisine}

\end{document}